\documentclass{ifacconf}

\usepackage{graphicx}      
\usepackage{natbib}   
\usepackage{amsmath}
\usepackage{amssymb}
\usepackage{booktabs}
\usepackage{tabularx}
\usepackage{subfig}
\usepackage{color}
\usepackage{soul}
\usepackage{xcolor}

\usepackage{draftwatermark}
\SetWatermarkText{ Accepted at  IFAC Workshop on Cyber-Physical Human Systems (CPHS 2026)}
\SetWatermarkScale{1.0}

\usepackage[final]{changes}

\begin{document}
\begin{frontmatter}

\title{Reason-guided Human Reward Modeling for Preference-based Reinforcement Learning \deleted{(PbRL)}} 

\thanks[footnoteinfo]{This research was supported, in part, by Cooperative Agreement W56HZV-24-2-0001 U.S. Army DEVCOM Ground Vehicle Systems Center. 
}

\author[First]{Anshul Nayak},
\author[First]{Shahil Shaik},
\author[Second]{Jonathon Smereka},
\author[First]{Yue Wang}

\address[First]{Clemson University,
   Clemson, SC 29631 USA (e-mail: anshuln@clemson.edu;
   shahils@clemson.edu;
   yue6@clemson.edu).}

\address[Second]{U.S. Army DEVCOM Ground Vehicle Systems Center,
   Warren, MI 48092 USA (e-mail: jonathon.m.smereka.civ@army.mil).}

\begin{abstract}
Reinforcement learning (RL) policies optimized for task rewards may produce behaviors that are efficient but misaligned with human expectations, safety preferences, or intent. This paper presents a \added[id=rev]{reason-guided} preference-based reinforcement learning (PbRL) framework that learns a human reward model from trajectory-level feedback. In addition to preference labels, the framework incorporates reason tags explaining why a behavior is preferred, enabling the learned reward to capture both human choices and their underlying rationale. Compared with preference-only reward learning, the reason-guided formulation provides additional supervision for learning human-aligned rewards and improves the interpretability of the resulting policy. The learned human reward is combined with the environment reward during policy optimization, allowing the agent to improve task performance while producing more interpretable, human-aligned, and preference-sensitive behavior. 

\end{abstract}


\end{frontmatter}

\section{Introduction}

Human preference plays a pivotal role in human-robot collaboration (\cite{WANG2025103497}). A robot optimized only for task performance may generate behaviors that are efficient from the robot's perspective but difficult for humans to trust or accept. For example, in social navigation, a robot may minimize travel time by passing close to obstacles or moving aggressively near humans, even though such behavior may appear unsafe or uncomfortable. In these scenarios, the quality of a decision should not be determined only by task performance, but also by whether the behavior aligns with human expectations, safety concerns, and preferences.

Classical approaches to human-aware decision-making have used optimal control, probabilistic inference, and temporal models 
to account for uncertainty in human-robot interaction (HRI). In particular, Hidden Markov Models (HMM), 
Dynamic Bayesian Networks (DBN) (\cite{xu2015optimo}), and Trust-
POMDPs (\cite{chen2018planning}) have been used to model latent human states such as intent, workload, or trust, and to update robot decision-making based on observations of human behavior or feedback. Although these models provide explicit uncertainty modeling, they often require carefully designed transition models, observation models, and reward functions. Online adaptation can also become challenging in high-dimensional or long-horizon settings, where repeated human queries and online re-planning may increase cognitive load and computational cost.

Alternatively, learning from human feedback has emerged as an effective alternative for aligning autonomous agents with human-preferred decisions (\cite{poddar2024personalizing}). In robotics, preference-based reinforcement learning (PbRL) allows humans to evaluate robot behaviors at the trajectory level and provide feedback that can be used to learn a reward model (\cite{liu2023efficient}). Reinforcement Learning from Human Feedback (RLHF) extends this idea by incorporating the learned human reward during policy optimization, enabling the robot to optimize not only for task completion but also for human-preferred behavior (\cite{liu2023efficient}). This is particularly important in HRI, where the environment reward may specify what the robot should accomplish, while human feedback provides information about how the task should be accomplished.

However, preference labels alone provide limited information about the factors that drive human decisions. A human may prefer one trajectory over another because it maintains safer obstacle clearance, follows a smoother path, avoids low-traversability regions, or better balances efficiency and safety. Standard preference-based reward learning captures which behavior is preferred, but does not explicitly model why it is preferred (\cite{wang2026preference}). Prior work has explored interpretable PbRL through structured reward representations, such as tree-based reward functions (\cite{bewley2021interpretable}), while broader studies on human-AI interpretability emphasize the importance of explaining agent behavior in terms that align with human expectations (\cite{sreedharan2021unifying}). However, standard preference-based reward learning typically does not use human-provided rationale as direct supervision for reward learning.

To address this limitation, we propose a reason-guided preference-based RLHF framework for human-aware robotic decision-making. In addition to trajectory preference labels, humans provide reason tags that explain why a behavior is preferred, such as predictability, legibility, safety, smoothness, or task efficiency. These reason labels provide auxiliary supervision during offline reward learning, improving the interpretability of the learned human reward model and supporting downstream policy optimization. The learned human reward is then combined with the task reward during reinforcement learning, enabling the robot policy to better balance task performance, safety, and human-aligned behavior.

The main contributions of this paper are as follows:
\begin{itemize}
\item We formulate a preference-based RLHF framework for human-aware robotic decision-making, where trajectory level human feedback is used to learn an offline reward signal that complements the task reward during policy optimization.
\item We develop a reason-guided human reward model that jointly predicts trajectory preferences and reason tags, enabling the robot to learn both which behaviors humans prefer and why they are preferred.
\item We show that incorporating reason-based feedback provides richer supervision than preference-only reward learning, improving reward interpretability and supporting policies that better balance task performance, safety, and human-aligned behavior.
\end{itemize}

\section{Methods}

\begin{figure*}[ht]
    \centering
    \includegraphics[width=0.75\linewidth]{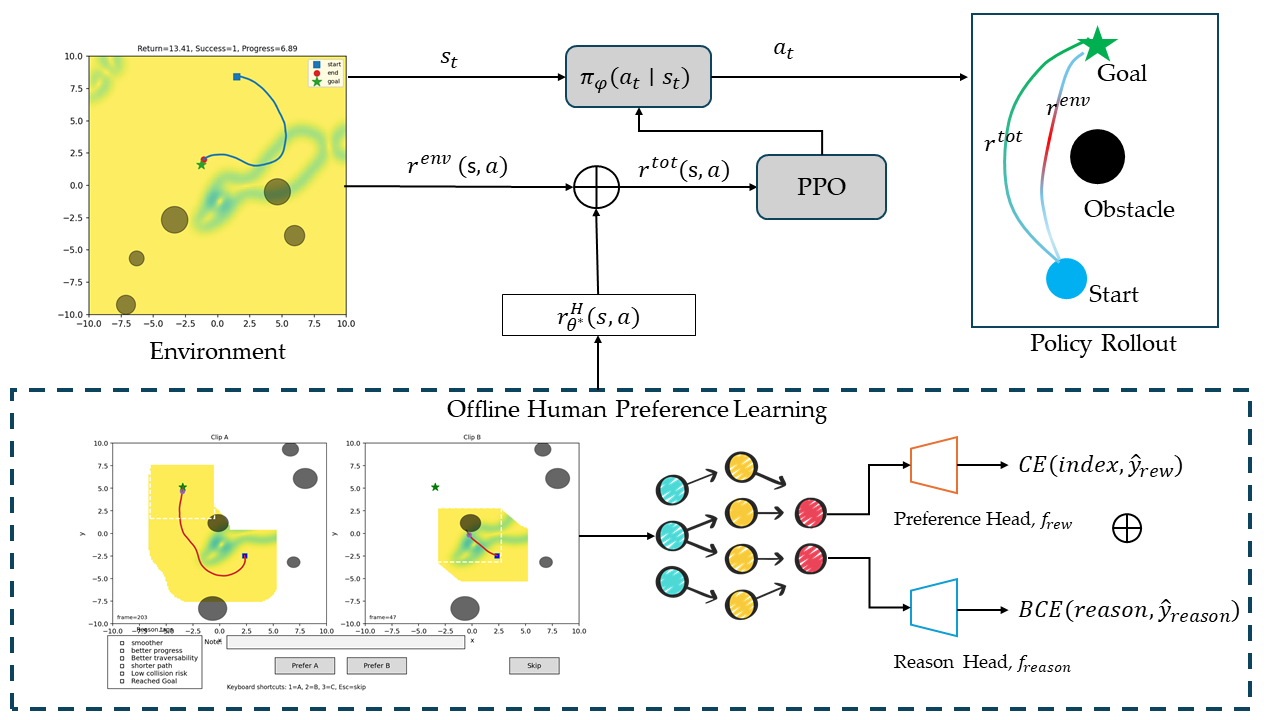}
    \caption{\textbf{ Offline  Human Preference Learning:} Human observes a pair of clips and records preference as well as reason labels. NN model is \added{trained offline to predict preference index $f^{rew}$ using cross entropy (CE) and reason tags $f^{reason}$ using binary cross entropy (BCE).} \textbf{Policy Optimization:} Learned human reward and task reward jointly perform policy optimization that is both task-efficient and aligned with interpretable human feedback. 
    }
    \label{fig:architecture}
\end{figure*}

\subsection{Problem Statement}

An RL agent must learn to leverage human preferences or  feedback to improve the policy so that the agent prefers human-aligned policy. We consider the episodic agent navigating in an environment as a MDP $\mathcal{M} = \{\mathcal{S}, \mathcal{A}, \mathcal{P}, r, \gamma\}$ where
    $s_t \in \mathcal{S}$ is the agent state and   $a_t \in \mathcal{A}$ is a continuous action with transition probability $s_{t+1} \sim  P(\cdot \,| s_t, \, a_t)$. The environment gives reward 
$r_{t}^{env} = r^{env}(s_t, a_t, s_{t+1})$. However, in preference-based RLHF, human might prefer an action aligned with \deleted{his/her} \added{their} own belief and preferences contrary to an optimal policy seeking safety over performance.  Therefore, we design the  reward as a combination of  learned human model and environment reward per step. 
\begin{equation}
    r_t^{\pi} =  r_{t}^{env} + \lambda_{H} \hat{r}_{t}^{H}
\end{equation}
where $\lambda_{H}$ controls the human preference strength and \added[id=rev]{$\hat{r}_{t}^{H}$ represents the predicted human scalar reward at time t.}
The goal is to learn a policy $\pi_{\varphi}(a_t \mid s_t)$ with the objective of maximizing a joint objective of task performance and human aligned preference.
\begin{equation}
    \pi^{*} = \underset{\pi}{\operatorname{argmax}} \,\,\mathbb{E}_{\tau \sim \pi} \left[\sum_{t=0}^{T} \gamma^t r_t^{\pi} \right]
\end{equation}

\subsection{Preference-based RLHF}

Preference-based RLHF uses trajectory-level human feedback to train a human reward model offline. A trajectory clip is a finite sequence of state-action transitions
$
    \tau = (s_0,a_0,s_1,a_1,\cdots,s_T),
$
where $T$ is the clip horizon length. For each transition, we construct an observable feature vector
$
    x_t = \phi(s_t,a_t,s_{t+1}) \in \mathbb{R}^{d},
$
where $\phi(\cdot)$ is a feature extraction function. The feature vector $x_t$ represents task- and safety-relevant quantities observable by the agent, such as progress, traversability, distance to goal, minimum collision distance, and other domain-specific features. For a  clip $\tau$, the corresponding sequence of features are
\begin{equation}
    \mathbf{X}(\tau)
    =
    [x_0 \mid\mid x_1 \mid\mid \cdots \mid\mid x_{T-1}]
    \in \mathbb{R}^{T \times d},
\end{equation}
where $\mid\mid$ denotes concatenation along the temporal dimension.
The human reward model is parameterized by a neural network with trainable parameters $\theta$. It maps the per-step feature vector $x_t$ to a scalar human reward
$
    r_{\theta}^{H}(x_t): \mathbb{R}^{d} \rightarrow \mathbb{R}.
$
A simple neural reward model can be written as
\begin{equation}
    h_t = f_{\theta}^{enc}(x_t),
\end{equation}
\begin{equation}
    \hat{r}_{\theta}^{H}(x_t) = f_{\theta}^{rew}(h_t),
\end{equation}
where $f_{\theta}^{enc}(\cdot)$ denotes the feature encoder and $f_{\theta}^{rew}(\cdot)$ denotes the scalar human reward head. 

In addition to predicting the scalar per-step human reward, the model also predicts the reason labels associated with the preferred trajectory clip. For this, a reason prediction head is added on top of the aggregated trajectory representation:
\begin{equation}
    \hat{\mathbf{c}}_{}(\tau)
    =
    \sigma\left(f_{\theta}^{reason}(\bar{h}_{\tau})\right),
\end{equation}
  \added[]{where $\sigma(\cdot)$ is the logistic sigmoid function,}  $\hat{\mathbf{c}}_{}(\tau) \in [0,1]^K$ is the predicted probability vector over $K$ reason labels, $f_{\theta}^{reason}(\cdot)$ is the reason classification head, and $\bar{h}_{\tau}$ is obtained by aggregating the per-step embeddings:
\begin{equation}
    \bar{h}_{\tau}
    =
    \frac{1}{T}
    \sum_{t=0}^{T-1}
    h_t .
\end{equation}

Let the trajectory clip be denoted by $\tau$ and the corresponding sequence of clip features be $\mathbf{X}(\tau)$. The predicted human reward for the full trajectory clip is computed by aggregating the per-step reward predictions:
\begin{equation}
\hat{R}_{\theta}^{H}(\mathbf{X}(\tau))
=
\sum_{t=0}^{T-1} \gamma^{t} \hat{r}_{\theta}^{H}(x_t),
\label{eq:clip_human_reward}
\end{equation}
where $\gamma \in [0,1]$ is a discount factor. 

\deleted[]{The human provides 
pairwise or index-based preference 
  comparing between  trajectory  rollouts. Additionally, the human also provides multiple reason labels for the \deleted[]{most-}preferred trajectory clip \deleted{over others.}} In the current preference-based reward learning, humans compare trajectory
clips generated under identical environmental conditions,
including start and goal locations and obstacle configurations.
For each pair $\{\tau_i^A,\tau_i^B\}$, where $i$ indexes an
environment configuration, the human selects the preferred
trajectory based on the agent's behavior. The reward model
$f_{\theta}^{\mathrm{rew}}(\cdot)$ is then trained to align
its predictions with these preferences. The human preference index is represented as 
\begin{equation} y_i = \begin{cases} 1, & \text{if } \tau_i^A \succ \tau_i^B, \\ 0, & \text{if } \tau_i^B \succ \tau_i^A. \end{cases} \end{equation} \added{where $\tau_i^A \succ \tau_i^B$ implies trajectory A is preferred over B.} Let $\mathbf{c}_i \in \{0,1\}^{K}$ denote the multi-hot encoded
reason vector for the human-preferred trajectory in the $i$-th
comparison, where $K$ is the number of predefined reasons.
\added{Each entry $c_{i,k}=1$ indicates the selection of reason $k$
($0$ otherwise), allowing multiple reasons to be selected
simultaneously for a single preference.} The reward model assigns each clip a scalar score $R_A = \hat{R}_{\theta}^{H}(\tau_i^A)$ and $R_B = \hat{R}_{\theta}^{H}(\tau_i^B)$. The probability that the human prefers trajectory clip $\tau_i^A$ over $\tau_i^B$ is \added{modeled using the Bradley-Terry preference model} (\cite{rafailov2023direct})  as
\begin{equation} P_{\theta}(\tau_i^A \succ \tau_i^B) = \sigma(R_A - R_B), \label{eq:preference_probability} \end{equation} 
The reward model is trained by minimizing the  cross-entropy (CE) loss over the preference dataset $\mathcal{D}_{H} = \{(\tau_i^A,\tau_i^B,y_i,\mathbf{c}_i)\}_{i=1}^{N}$. The preference loss using the reward difference as the logit is represented as \begin{equation} \begin{split} \mathcal{L}_{pref}(\theta) = -\frac{1}{N} \sum_{i=1}^{N} \Big[ y_i \log \sigma(R_A-R_B) +\\ (1-y_i)\log \sigma(R_B-R_A) \Big]. \end{split} \end{equation}

Preference comparisons constrain only relative ordering between trajectory clips and do not uniquely identify the per-step reward function. Hence, we regularize the learned reward to reduce large reward magnitudes and high temporal variation before using it for policy optimization.

\begin{equation}
\begin{split}
\mathcal{L}_{reg}(\theta)
=
\frac{1}{N}
\sum_{i=1}^{N}
\sum_{\tau \in \{\tau_i^A,\tau_i^B\}}
\Bigg[
\frac{1}{T}
\sum_{t=0}^{T-1}
\left(r_{\theta}^{H}(x_t)\right)^2
+\\
\frac{1}{T-1}
\sum_{t=0}^{T-2}
\left(
r_{\theta}^{H}(x_{t+1})
-
r_{\theta}^{H}(x_t)
\right)^2
\Bigg].
\end{split}
\end{equation}
The first term controls reward magnitude, while the second encourages temporal smoothness. 

In addition, reason labels provide semantic supervision for reward learning. For the preferred clip, the predicted reason vector is
\begin{equation}
\hat{\mathbf{c}}_{i}
=
y_i \hat{\mathbf{c}}_{}(\tau_i^A)
+
(1-y_i)\hat{\mathbf{c}}_{}(\tau_i^B).
\end{equation}
The reason-guided binary cross-entropy (BCE) loss is defined as
\begin{equation}
\begin{split}
\mathcal{L}_{reason}(\theta)
=
-\frac{1}{N}
\sum_{i=1}^{N}
\sum_{k=1}^{K}
\Big[
c_{i,k}\log \hat{c}_{i,k}
+\\
(1-c_{i,k})\log(1-\hat{c}_{i,k})
\Big].
\end{split}
\end{equation}
This loss encourages the reward model to capture not only which trajectory is preferred, but also why it is preferred. 
The final offline reward-model training objective is
\begin{equation}
\mathcal{L}_{H}(\theta)
=
\mathcal{L}_{pref}(\theta)
+
\alpha_{reason}\mathcal{L}_{reason}(\theta)
+
\beta_{reg}\mathcal{L}_{reg}(\theta),
\end{equation}
where $\alpha_{reason}$ and $\beta_{reg}$ are weights for the reason and regularization loss terms, respectively.


\subsection{Policy Optimization}
The trained model \added{$\hat{r}_{\theta^{*}}^{H}$} is then used as a learned human reward function during agent learning where the human reward model parameters $\theta^{*}$ are frozen. At each time step, the  reward is queried per-step for policy optimization as
\begin{equation}
r_t^{total}
=
\tilde{r_t}^{env}
+
\lambda_H \tilde{r}_{\theta^{*}}^{H}(x_t),
\label{eq:total_reward}
\end{equation}

where, $\tilde{r}$ represents normalized  per-step scalar reward. The policy $\pi_{\varphi}(a_t\mid s_t)$  is trained to maximize the objective:
\begin{equation}
    J(\varphi) = \mathbb{E}_{\tau \sim \pi_{\varphi}}[\sum_{t=0}^{T} \gamma^t(\tilde{r_t}^{env}
+
\lambda_H \tilde{r}_{\theta^{*}}^{H}(x_t))]
\end{equation}  where \added{$J(\varphi)$ represents the expected discounted return and}  the \added{trainable policy parameters $\varphi$ } are optimized using gradient descent $\nabla_\varphi J(\varphi)$.



\section{Experiments}

\subsection{Data Collection}

We designed an off-road single-agent navigation simulator. The overall task for the agent is to navigate to the goal $(x_g, y_g)$ while adhering to robot dynamics and obstacles. The agent can navigate using pre-trained RL policy or model-based control approaches. The robot has stochastic dynamics and the robot pose $[x, y, \theta]^{T}$ evolves under control command $[\omega, v]$. The environment also contains static obstacles with varying sizes and configuration to induce collision avoidance.  \deleted{Additionally, to simulate off-road conditions, terrain is generated in Isaacsim using Perlin-
style noise and  random rough patches.  The local traversability is computed from gradients and robot actuator limits confined to a score between $[0, 1]$}. \comment[]{No Terrain Traversability}

Pre-trained RL policy or model-based planners are used to create the offline preference dataset. The dataset $\mathcal{D}$ contains multiple sampled clips along with per-step  features $x_{t}$ for the same start and goal locations and obstacle configurations. The human is shown pair-wise episodic clips $\{\tau_A, \tau_B\}$ and needs to select the preferred trajectory along with reason labels for the selection.  For, a simple go-To-goal environment with obstacle avoidance, the reason labels for preferred trajectory can be due to "safe clearance", "shorter path", "smoothness" 
and so on. The dataset stores the preferred clip index $\{0, 1\}$ along with the reason labels represented as multi-hot encoded vector    for each query.  

\subsection{Training Implementation}

A simple MLP with one hidden layer maps the clip-level input features ${\mathbf{X}(\tau)}$ into embedded representation. The output contains two heads, a scalar reward head, $f_\theta^{rew}$ that minimizes the cross entropy (CE) loss between predicted  and actual human preferred index as well as a reason head $f_\theta^{reason}$ that minimizes binary cross entropy (BCE) loss across multiple labels simultaneously (Fig.\ref{fig:architecture}). The dataset $\mathcal{D}$ contains dedicated human annotation across 350  queries with 90\% queries used for training and 10\% for validation. The parameters of the trained human reward model $\hat{R}_{\theta^{*}}^{H}$ are frozen and queried at step-level during policy training. The training hyperparameters are shown in Table.\ref{tab:hyperparameters}  

\subsection{Policy Training }

The policy $\pi_{\varphi} (a_t \mid s_t)$ is trained on a vectorized 2D off-road environment for go-To-goal task with obstacle avoidance. The \added{normalized} task reward $\tilde{r}^{env}$ is designed to encourage progress toward the goal while penalizing unsafe or inefficient behavior. The environment provides a large positive terminal reward for reaching the goal and large negative terminal penalties for collisions or going out-of bounds. Reward shaping  include progress towards  goal,  obstacle-proximity and control-related penalties for inducing trajectory smoothness. Adding human reward $\tilde{r}_{\theta^{*}}^{H}$ and training the policy with $r^{total}$ (\ref{eq:total_reward}) is expected to align the agent's decision with human's preference. 

\begin{table}[ht]
\centering
\caption{\added{Hyperparameters}}
\label{tab:hyperparameters}

\small
\renewcommand{\arraystretch}{0.85}
\setlength{\tabcolsep}{4pt}

\begin{tabularx}{\columnwidth}{@{}X l@{}}
\toprule
\textbf{Parameter} & \textbf{Value} \\ 
\midrule
Hidden layer dim & 256 \\
Optimizer & SGD \\
Learning Rate & $2 \times 10^{-4}$ \\
Effective Batch Size & 8 \\
epochs &20 \\
$\lambda_H$ & $[0, 0.5]$ \\
$\gamma$ & $0.9$ \\
$\alpha_{reason}$ & $0.2$ \\
$\beta_{reg}$ & $1e-3$ \\
\bottomrule
\end{tabularx}

\end{table}

\subsection{Reason-guided Ablation} \label{sec:reason_guided_ablation}

To isolate the contribution of reason supervision and reward regularization, we train the human reward model offline under four ablation objectives: preference loss ($\mathcal{L}_{pref}$) only, preference loss with reward regularization ($\mathcal{L}_{\mathrm{pref}}+\mathcal{L}_{\mathrm{reg}}$), preference loss with reason supervision ($\mathcal{L}_{\mathrm{pref}}+\alpha_{reason}\mathcal{L}_{\mathrm{reason}}$), and preference loss with both reason supervision and reward regularization ($\mathcal{L}_{\mathrm{pref}}+\alpha_{\mathrm{reason}}\mathcal{L}_{\mathrm{reason}}+\mathcal{L}_{\mathrm{reg}}$). After training, each reward model is saved and kept frozen during policy optimization. The downstream policy is then trained using the combined environment and learned human reward, where the human reward contribution is scaled by $\lambda_h \in \{0.1,0.2\}$. This design separates reward-model learning from policy optimization and allows us to evaluate whether reason-guided and regularized reward models produce policies with better task performance, obstacle clearance, path efficiency, and success rate \added{(agent avoided collisions and reached goal)} compared with the environment-only simple preference ($\mathcal{L}_{pref}$) baseline.

\begin{figure}[t]
\centering

    \subfloat[\centering safe clearance to obstacles]{{\includegraphics[width=5cm]{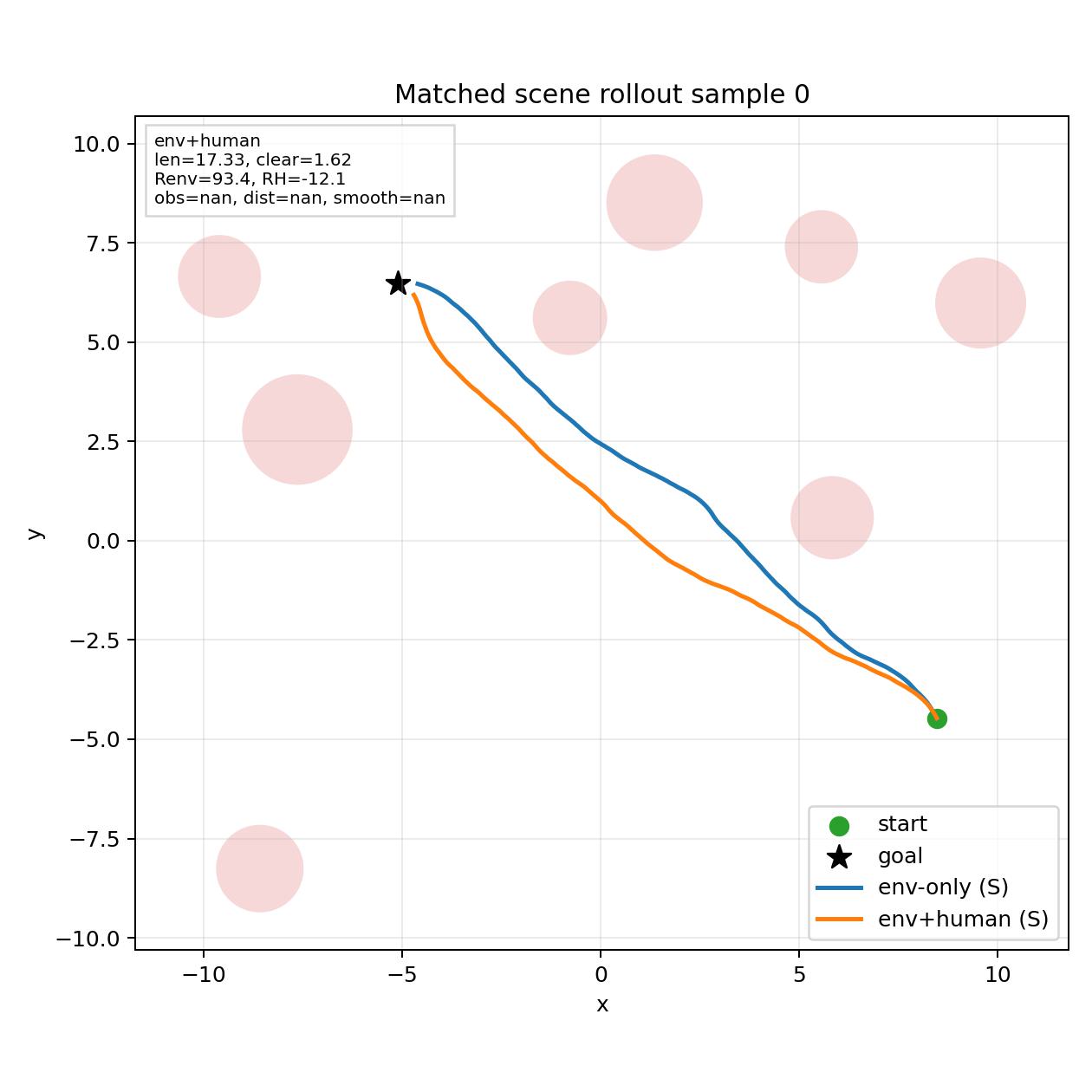} }
    \label{fig:safe_clearance}}%
    
    \subfloat[\centering smoother and shorter path]{{\includegraphics[width=5cm]{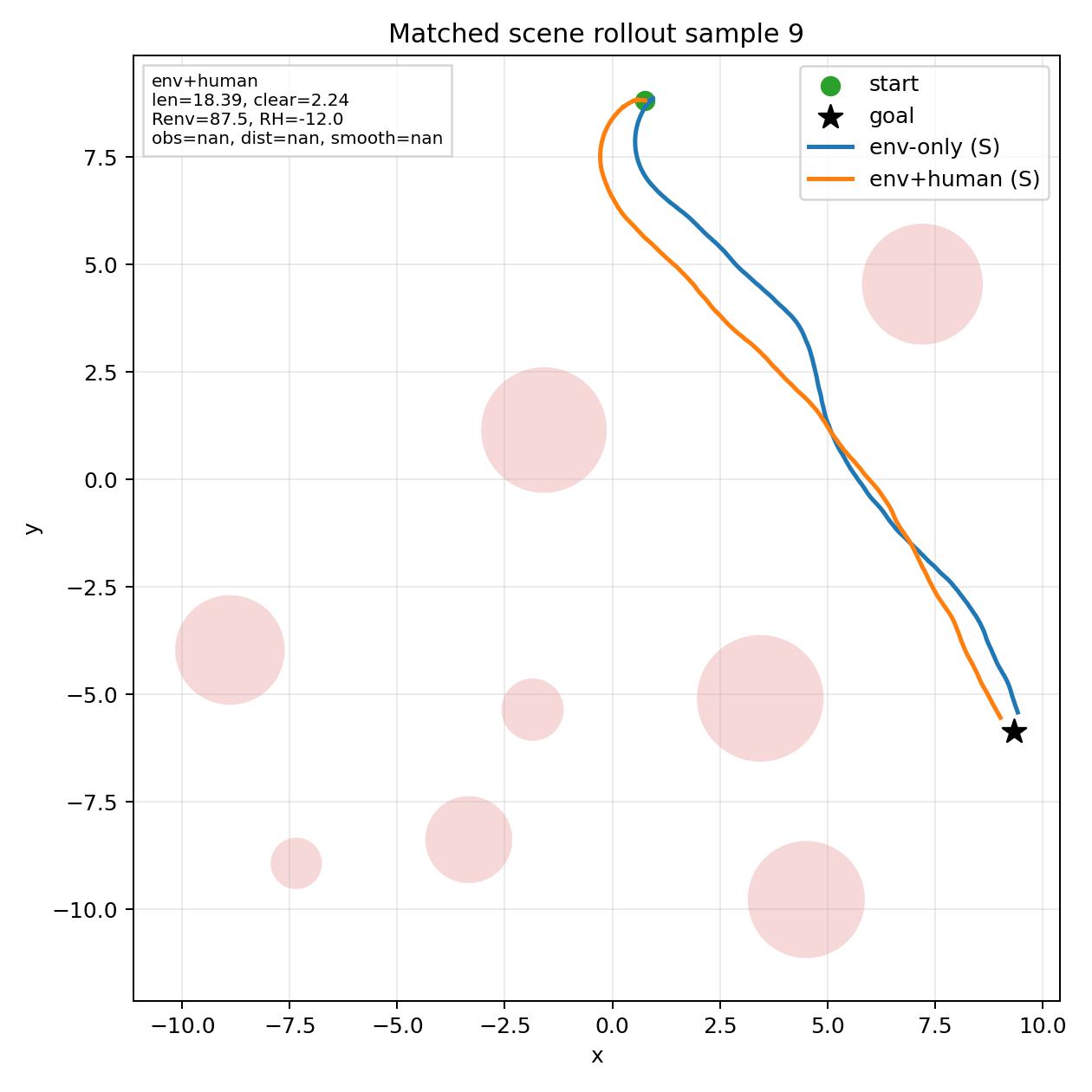} }
    \label{fig:smooth_short}}%
    \caption{Policy Evaluation Comparison: successful rollout comparisons between the environment-only  and the combined environment and human policy under similar environment conditions.}%
    %


\label{fig:rollout_comparison}
\end{figure}

\section{Results}
\label{sec:results}

This section evaluates whether incorporating the learned human reward improves the  alignment of the trained policy towards human preferences when compared against a policy optimized using only the task reward. Initially, we compare two reward settings: (i) an environment-only reward baseline, denoted as $R_{\mathrm{env}}$, and (ii) a combined reward formulation, denoted as $R_{\mathrm{env}} + \lambda_H \,\hat{R}_{\theta}^{H}$, where $\hat{R}_{\theta}^{H}$ is obtained by querying the learned human feedback model. Secondly, we also highlight the importance of reason-guided reward  training (sec. \ref{sec:reason_guided_ablation}) and regularization over simple preference based training. We evaluate the policies under the same environment configuration  comparing  human reward model trained only using preference loss, preference and reason loss and finally, regularized preference and reason loss. We analyze policy rollouts across multiple scene and policy seeds using task-level and safety-related trajectory metrics. 

\subsection{Rollout-Based Policy Comparison}
\label{subsec:rollout_results}

To compare the behavior induced by the two reward formulations, we evaluate the trained policies using a fixed rollout protocol. For each method, we generate $30$ policy rollouts across $4$ scene seeds and multiple policy seeds, using the same start and goal conditions along with obstacle configuration  to ensure a fair comparison. This setup allows us to compare the behavior of the task-only policy and the human-reward-augmented policy under same environmental conditions.

\begin{table}[h!]
\centering
\caption{Reward metrics for rollout evaluation at $\lambda_H=0.2$.}
\label{tab:rollout_metrics}
\resizebox{0.95\columnwidth}{!}{%
\begin{tabular}{lcccc}
\hline
Policy   & Path length & Success (\%) \\
\hline
$R_{\mathrm{env}}$ 
 & $11.02 \pm 0.96$ & $67.5 \pm 4.2$ \\
$R_{\mathrm{env}}+\lambda_H R_{\mathrm{human}}$ 
& $10.92 \pm 0.19$ & $77.5 \pm 5.7$ \\
\hline
\end{tabular}%
}
\end{table}

Figure~\ref{fig:rollout_comparison} shows representative trajectory rollouts for  two reward formulations, $R_{env}$ and $R_{tot} = R_{env} + \hat{R}_{\theta}^{H}$. Each scene has same environment configuration and only successful rollouts are considered for both the policies such that the agent reaches goal to effectively capture human preference. In Fig.\ref{fig:safe_clearance}, the environment-only policy generates a path that is much closer to the obstacle while the  policy trained with $R_{tot}$ generates a smoother path while maintaining safe clearance with obstacles. This aligns with human preference data where the human preferred paths that are  "shorter", "smoother" and had "safer clearance" from obstacles when selecting among a pair of trajectories. Similarly, Fig.\ref{fig:smooth_short} shows the combined human and environment reward model prefers a shorter and smoother path while reaching the goal. Since, the environment reward does not explicitly encode human preference, it  may yield
trajectories closer to obstacles, while combined reward
$R_{\mathrm{env}} + \lambda_H \hat{R}_{\theta^*}^{H}$
encourages trajectories that better align with human
preferences for safer obstacle clearance and smoother motion.

We quantitatively evaluate each policy over four held-out scene seeds and report the mean and standard deviation of trajectory-level metrics. The table summarizes combined reward, obstacle clearance, path length, and task success rate. Combined reward measures the total rollout return under the evaluation reward, while obstacle clearance captures the minimum separation from obstacles or unsafe regions along the trajectory. Path length measures navigation efficiency, and success rate reports the percentage of rollouts in which the robot reaches the goal without collision within the allowed horizon.  The learned human reward policy achieves a shorter average path length while maintaining a higher success rate, suggesting more efficient and smoother trajectories aligning with human preferences. \deleted{Although obstacle clearance is slightly lower for $\lambda_H=0.2$, this appears to be driven by a small number of environment-only rollouts that remain unusually far from obstacles, which can skew the mean when averaging over only four scene seeds.}

\begin{table*}[t]
\centering
\caption{Ablation study comparing reward-model training objective. 
 Preference-only baseline $\mathcal{L}_{\mathrm{pref}}$, while environment+human rows use the corresponding ablation objective.}
\label{tab:reward_ablation_results}
\scriptsize
\begin{tabular}{llccccc}
\toprule
$\lambda_H$  & Training objective 
& Obstacle clearance 
& Path length 
& Success rate (\%) 
& subjective Evaluation (\%) \\
\midrule
0.1 
& $\mathcal{L}_{\mathrm{pref}}$
& $1.82 \pm 0.20$
& $11.44 \pm 1.46$
& $66.7 \pm 4.7$ 
& $33.1 \pm 6.4$ \\

0.1 
& $\mathcal{L}_{\mathrm{pref}}+\alpha_{\mathrm{reason}}\mathcal{L}_{\mathrm{reason}}$
& $1.90 \pm 0.27$
& $11.25 \pm 0.52$
& $73.3 \pm 0.0$ 
& $65.4 \pm 7.2$ \\

0.1 
& $\mathcal{L}_{\mathrm{pref}}+\mathcal{L}_{\mathrm{reg}}$
& $1.80 \pm 0.23$
& $11.16 \pm 0.36$
& $75.0 \pm 2.4$ 
& $68.3 \pm 8.1$ \\

0.1 
& $\mathcal{L}_{\mathrm{pref}}+\alpha_{\mathrm{reason}}\mathcal{L}_{\mathrm{reason}}+\mathcal{L}_{\mathrm{reg}}$
& $\textbf{1.99} \pm 0.30$
& $11.29 \pm 0.87$
& $\textbf{78.3} \pm 2.4$ 
& $67.2 \pm 5.1$ \\

\midrule

0.2 
& $\mathcal{L}_{\mathrm{pref}}$
& $1.78 \pm 0.14$
& $11.72 \pm 1.33$
& $65.0 \pm 2.4$ 
& $23.1 \pm 4.9$ \\

0.2 
& $\mathcal{L}_{\mathrm{pref}}+\alpha_{\mathrm{reason}}\mathcal{L}_{\mathrm{reason}}$
& $2.01 \pm 0.14$
& $11.65 \pm 0.52$
& $68.3 \pm 2.4$ 
& $66.6 \pm 4.7$ \\

0.2 
& $\mathcal{L}_{\mathrm{pref}}+\mathcal{L}_{\mathrm{reg}}$
& $2.05 \pm 0.05$
& $12.04 \pm 1.11$
& $81.7 \pm 7.1$ 
& $81.8 \pm 4.1$ \\

0.2 
& $\mathcal{L}_{\mathrm{pref}}+\alpha_{\mathrm{reason}}\mathcal{L}_{\mathrm{reason}}+\mathcal{L}_{\mathrm{reg}}$
& $\textbf{2.18} \pm 0.07$
& $\textbf{11.20} \pm 1.33$
& $73.3 \pm 4.7$ 
& $80.0 \pm 4.7$ \\

\bottomrule
\end{tabular}
\end{table*}

\subsection{Reason-guided Reward Training}

\begin{figure}[t]
\centering

\subfloat[Training loss]{%
    \includegraphics[width=0.47\columnwidth]{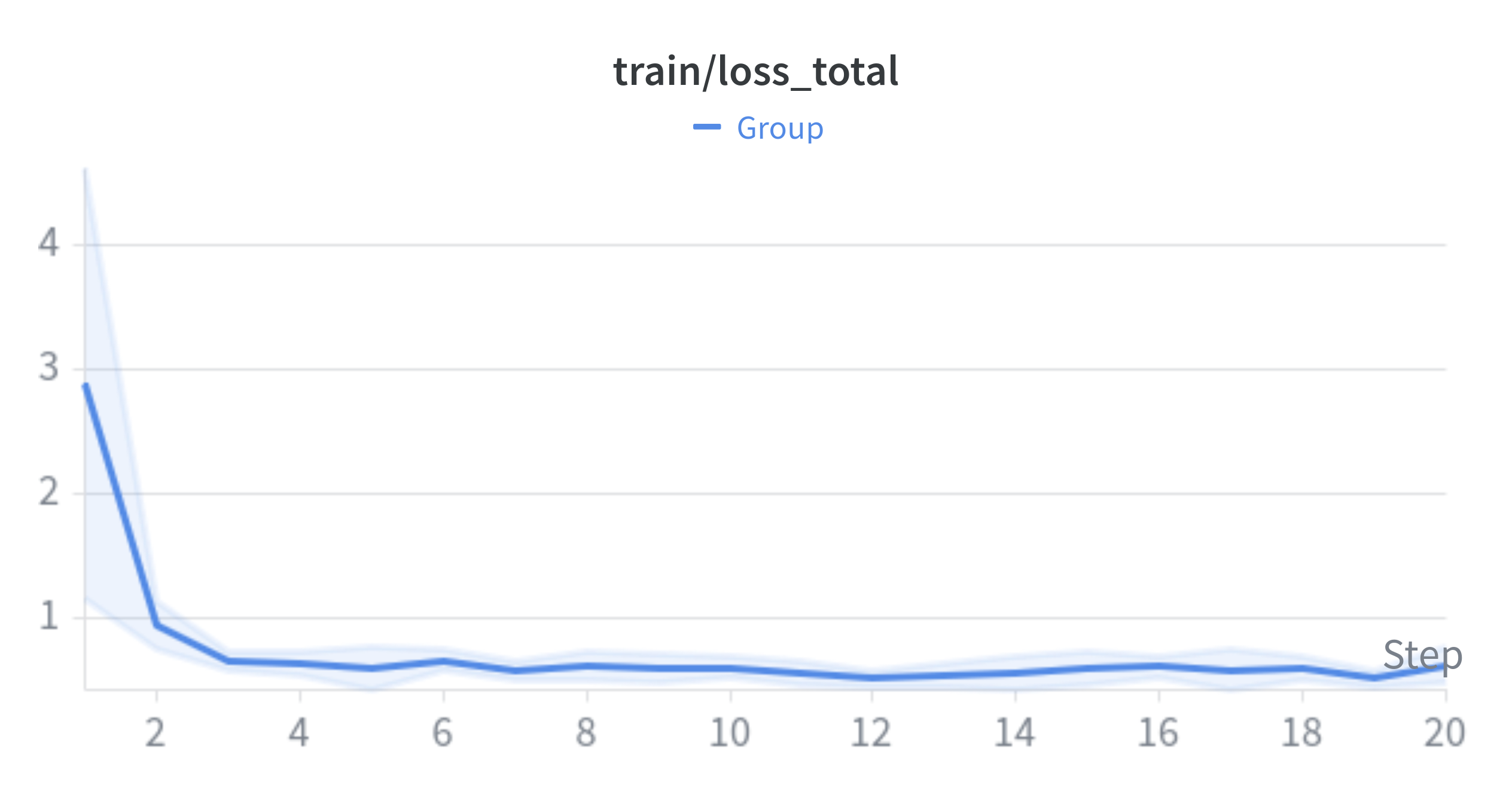}
    \label{fig:val_loss_curve}
}
\hfill
\subfloat[Validation accuracy]{%
    \includegraphics[width=0.47\columnwidth]{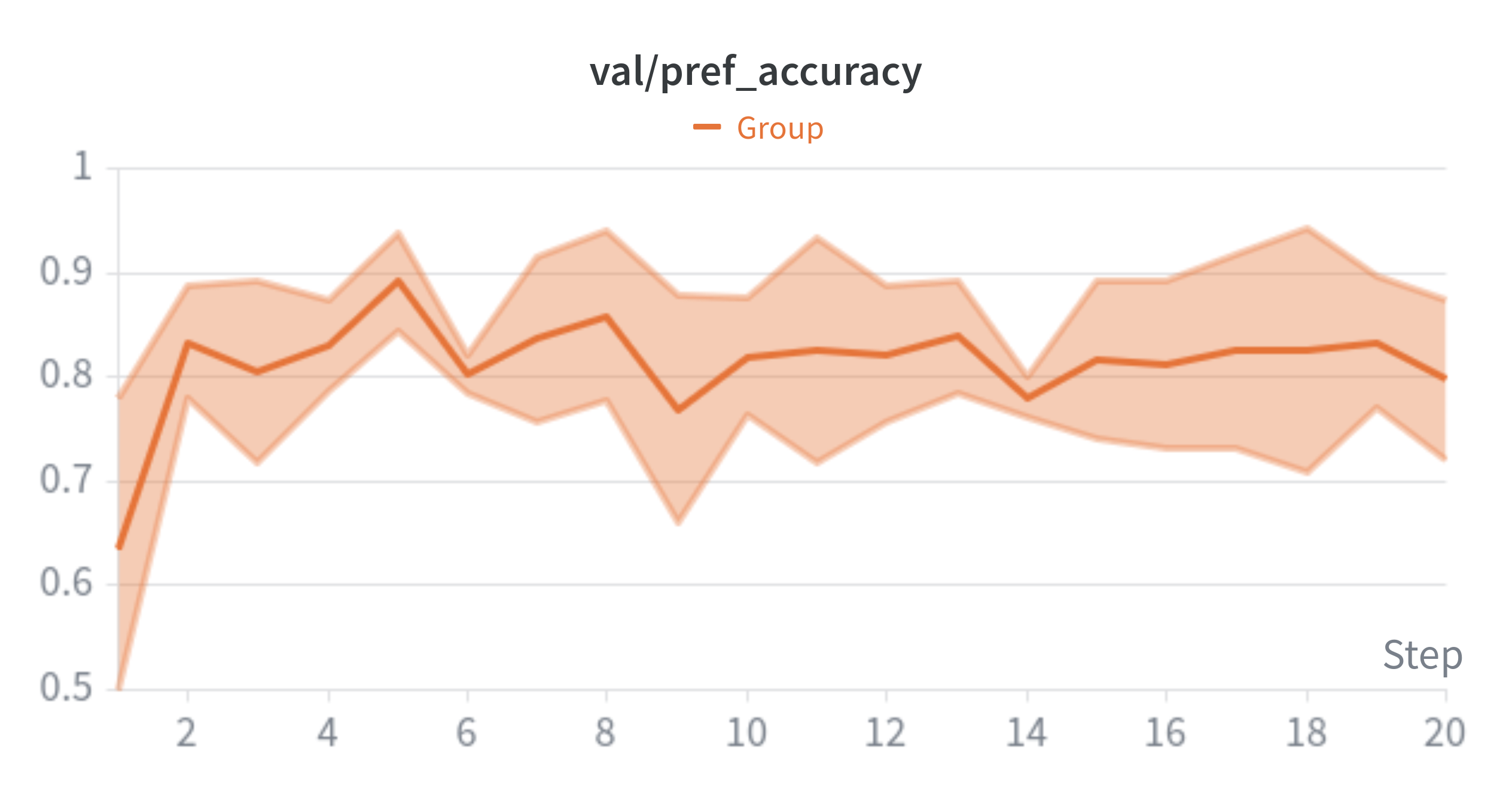}
    \label{fig:val_score_curve}
}

\vspace{0.4em}

\subfloat[Policy Training]{%
    \includegraphics[width=0.7\columnwidth]{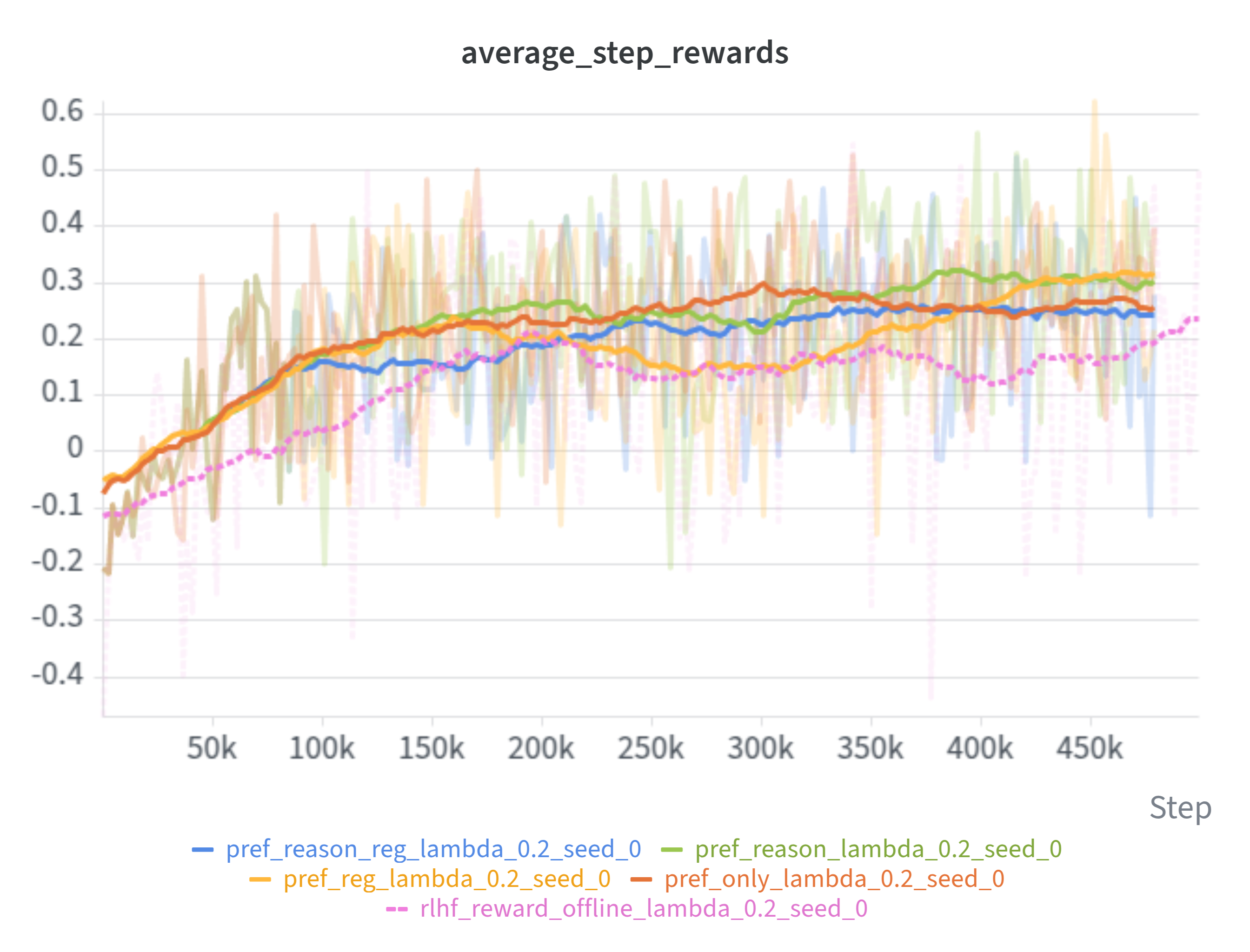}
    \label{fig:policy_training_curve}
}

\caption{Reward-model validation and policy-training curves for ablation variants.}
\label{fig:ablation_training_curves}
\end{figure}

\label{subsec:rollout_results}

Figure \ref{fig:ablation_training_curves} presents the reward-model training, validation, and
downstream policy-training curves across the ablation variants.
The decreasing training loss and improving validation accuracy
indicate stable reward learning across objectives. The
reason-guided variants incorporate additional semantic
supervision, while reward regularization constrains reward
magnitude and temporal variation without destabilizing training.

Table.\ref{tab:reward_ablation_results} evaluates the effect of these objectives on downstream
policy performance. At $\lambda_H=0.1$, combining reason
supervision and regularization improves task success from
$66.7 \pm 4.7\%$ to $78.3 \pm 2.4\%$ compared with
preference-only learning, while increasing obstacle clearance
from $1.82 \pm 0.20$ to $1.99 \pm 0.30$. At $\lambda_H=0.2$, the combined objective achieves the highest
obstacle clearance ($2.18 \pm 0.07$) with a path length of
$11.20 \pm 1.33$. However, regularization alone achieves the
highest success rate ($81.7 \pm 7.1\%$), highlighting the
trade-offs between safety and task completion. Overall,
the results suggest that reason supervision and regularization
can improve downstream policy behavior, although their
relative benefits depend on the human-reward weight and
evaluation metric.

\subsection{Blind subjective study}

Three annotators independently assessed matched trajectory pairs in a blinded evaluation. Trajectories were presented with neutral A/B labels with policy  randomized  such that policy identities and reward values hidden from annotator. Each annotator selected a preferred trajectory or skipped the comparison; responses were subsequently unblinded using the stored ground truth mappings. Preference rates were computed by aggregating choices across multiple runs and seeds within each annotator, then mean and sample standard deviation across the three annotators are reported in Table.\ref{tab:reward_ablation_results}. Adding reason supervision to the preference loss improved both task success and subjective evaluation: subjective scores increased from $33.1\pm6.4\%$ to $65.4\pm7.2\%$ and from $23.1\pm4.9\%$ to $66.6\pm4.7\%$ for lambdas $0.1$ and $0.2$ respectively. The $\mathcal{L}_{\mathrm{pref}}+\mathcal{L}_{\mathrm{reg}}$ objective also performed comparably to the full objective, achieving subjective scores of $68.3\pm8.1\%$ versus $67.2\pm5.1\%$ at $\lambda_H=0.1$, and $81.8\pm4.1\%$ versus $80.0\pm4.7\%$ at $\lambda_H=0.2$. These results support the benefit of reason supervision for learning rewards aligned with human preferences, while showing that preference learning with regularization provides a strong alternative to the full objective.

\section{Conclusion}

This paper presents a reason-guided preference-based RLHF framework for human-aware robotic decision-making. The proposed approach learns an offline human reward model from trajectory-level preference feedback and augments preference learning with reason labels that explain why a behavior is preferred. By jointly learning trajectory preferences and their associated rationale, the reward model provides a more interpretable representation of human-aligned behavior than preference-only reward learning. The learned human reward is then combined with the task reward during policy optimization, allowing the robot to account for both task performance and human-sensitive criteria such as safety, smoothness, predictability, and traversability. The results show that incorporating learned human feedback can influence policy behavior beyond what is obtained from the task reward alone. In particular, the reason-guided reward model provides additional supervision during offline reward learning and supports policies that better balance efficiency with human-aligned behavioral preferences. The ablation study further highlights the role of reason supervision and reward regularization in shaping the learned reward model and downstream policy performance.
Future work will extend the framework to  HRI settings with richer reward signals and trust-based reward for  multi-agent robotic teams.

\bibliography{ifacconf}             

@article{WANG2025103497,
title = {A reinforcement learning from human feedback based method for task allocation of human robot collaboration assembly considering human preference},
journal = {Advanced Engineering Informatics},
volume = {66},
pages = {103497},
year = {2025},
issn = {1474-0346},
doi = {https://doi.org/10.1016/j.aei.2025.103497},
url = {},
author = {Jingfei Wang and Yan Yan and Yaoguang Hu and Xiaonan Yang}
}

@article{poddar2024personalizing,
  title={Personalizing reinforcement learning from human feedback with variational preference learning},
  author={Poddar, Sriyash and Wan, Yanming and Ivison, Hamish and Gupta, Abhishek and Jaques, Natasha},
  journal={Advances in Neural Information Processing Systems},
  volume={37},
  pages={52516--52544},
  year={2024}
}

@inproceedings{xu2015optimo,
  title={Optimo: Online probabilistic trust inference model for asymmetric human-robot collaborations},
  author={Xu, Anqi and Dudek, Gregory},
  booktitle={Proceedings of the tenth annual ACM/IEEE international conference on human-robot interaction},
  pages={221--228},
  year={2015}
}

@inproceedings{chen2018planning,
  title={Planning with trust for human-robot collaboration},
  author={Chen, Min and Nikolaidis, Stefanos and Soh, Harold and Hsu, David and Srinivasa, Siddhartha},
  booktitle={Proceedings of the 2018 ACM/IEEE international conference on human-robot interaction},
  pages={307--315},
  year={2018}
}

@article{wang2026preference,
  title={Preference-based Reinforcement Learning for Robotics: Advancements and Open Challenges},
  author={Wang, Ruiqi},
  journal={Authorea Preprints},
  year={2026},
  publisher={Authorea}
}

@article{bewley2021interpretable,
  title={Interpretable preference-based reinforcement learning with tree-structured reward functions},
  author={Bewley, Tom and Lecue, Freddy},
  journal={arXiv preprint arXiv:2112.11230},
  year={2021}
}

@article{sreedharan2021unifying,
  title={A Unifying Bayesian Formulation of Measures of Interpretability in Human-AI},
  author={Sreedharan, Sarath and Kulkarni, Anagha and Smith, David E and Kambhampati, Subbarao},
  journal={arXiv preprint arXiv:2104.10743},
  year={2021}
}

@article{liu2023efficient,
  title={Efficient preference-based reinforcement learning using learned dynamics models},
  author={Liu, Yi and Datta, Gaurav and Novoseller, Ellen and Brown, Daniel S},
  journal={arXiv preprint arXiv:2301.04741},
  year={2023}
}

@inproceedings{rafailov2023direct,
  title     = {Direct Preference Optimization: Your Language Model is Secretly a Reward Model},
  author    = {Rafailov, Rafael and Sharma, Archit and Mitchell, Eric and Manning, Christopher D. and Ermon, Stefano and Finn, Chelsea},
  booktitle = {Advances in Neural Information Processing Systems (NeurIPS)},
  volume    = {36},
  pages     = {53728--53741},
  year      = {2023},
  url       = {}
}
                                                   







\end{document}